\documentclass[10pt,twocolumn,letterpaper]{article}

\usepackage[pagenumbers]{cvpr} 

\definecolor{cvprblue}{rgb}{0.21,0.49,0.74}
\usepackage[pagebackref,breaklinks,colorlinks,allcolors=cvprblue]{hyperref}

\def\confName{CVPR}
\def\confYear{2027}

\usepackage{xspace}
\usepackage{multirow}
\usepackage{makecell}
\usepackage{graphicx}

\newcommand{\bench}{MAVEBench\xspace}
\newcommand{\method}{MAVE\xspace}
\newcommand{\dataset}{MAVEData-3K\xspace}

\usepackage[table]{xcolor}
\usepackage{booktabs}
\usepackage{multirow}
\usepackage{graphicx}
\usepackage{pifont}

\definecolor{midpurblue}{RGB}{205,227,255}
\definecolor{lightpurblue}{RGB}{243,249,255}
\definecolor{lightred}{RGB}{254,240,236}
\definecolor{lightgreen}{RGB}{235,255,240}
\definecolor{refcolor}{gray}{0.9}

\newcommand{\best}[1]{\cellcolor{midpurblue}{#1}}
\newcommand{\secondbest}[1]{\cellcolor{lightpurblue}{#1}}

\newcommand{\bestab}[1]{\textbf{#1}}

\newcommand{\midsize}{\fontsize{9.5}{11.5}\selectfont}
\newcommand{\authorspace}{\quad}
\title{MAVE: Reconstructing Editable 3D Blockouts from Film Shots}

\author{
    Yifan Chang\textsuperscript{1,2,5,}\thanks{Equal contribution.}
    \authorspace
    Jian Yang\textsuperscript{1,3,*}
    \authorspace
    Jiangnan Shao\textsuperscript{1,4,*}
    \authorspace
    Xiaofeng Wang\textsuperscript{6}
    \\
    Zhaoxin Fan\textsuperscript{7}
    \authorspace
    Renjing Xu\textsuperscript{4}
    \authorspace
    Ming Lu\textsuperscript{8}
    \authorspace
    Xingang Wang\textsuperscript{2,}\thanks{Corresponding author.}
    \authorspace
    Zheng Zhu\textsuperscript{1,\dag}
    \vspace{2mm}
    \\
    {\midsize\textsuperscript{1}GigaAI} \quad
    {\midsize\textsuperscript{2}CASIA} \quad
    {\midsize\textsuperscript{3}RUC} \quad
    {\midsize\textsuperscript{4}HKUST (GZ)} \quad
    {\midsize\textsuperscript{5}UCAS} \quad
    {\midsize\textsuperscript{6}THU} \quad
    {\midsize\textsuperscript{7}BUAA} \quad
    {\midsize\textsuperscript{8}ZPark IPC}
    \\[1mm]
    {\midsize Project page: \href{https://yfchang-cv.github.io/MAVE/}{\texttt{yfchang-cv.github.io/MAVE}}}
}

\begin{document}
\maketitle
\begin{abstract}
Animated 3D blockouts support previsualization and provide structural conditions for video generation, but manually reconstructing them from reference videos is costly. This limits both their use by creators and the production of paired training data. We introduce \bench, a benchmark of 50 film shots and 416 keyframes for evaluating camera motion, spatial layout, and character blocking. We propose \method, a video-to-blockout framework combining \textbf{M}etric observation, \textbf{A}gentic modeling, and evidence-based \textbf{VE}rification. It first recovers cameras, human motion, and geometric evidence in a shared metric coordinate system. An agent then constructs an editable environment under constraints from these observations, while programmatic checks report inconsistencies to guide refinement. The outputs are editable Blender scenes and frame-aligned blockout videos. On \bench, \method improves overall scores by 27.5--41.1 points across three agents over their corresponding vanilla baselines. With DeepSeek-1017, it achieves a score of 73.2 at an average API cost of only \$0.054 per shot, outperforming the strongest vanilla baseline while reducing total reconstruction time by 53\%. We also construct \dataset, a dataset of 3,000 blockout--video pairs for model training, and will release it publicly. Creators can change viewpoints and character paths, then render the edited scenes to condition video generation. \footnote{Code is available. \url{https://github.com/yfChang-cv/MAVE}}
\end{abstract}    
\vspace{-8pt}
\section{Introduction}
\label{sec:intro}
\begin{figure}[t]
    \centering
    \includegraphics[width=\columnwidth]{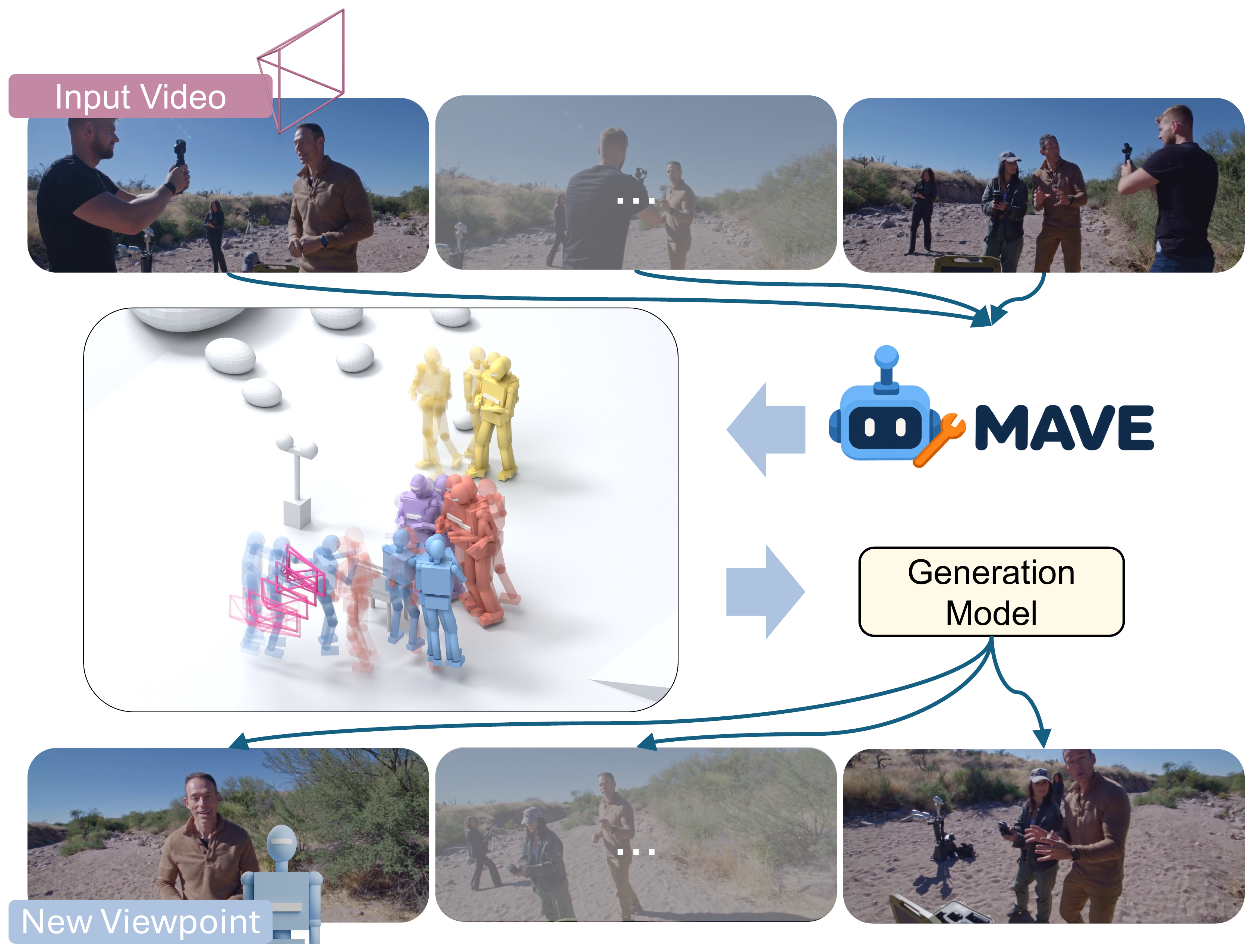}
    \caption{\textbf{From reference video to editable 3D blockout.}
    \method reconstructs scene geometry, character motion, and the camera trajectory from an input video. The resulting blockout can be edited and rendered from a new viewpoint to guide video generation.}
    \label{fig:first_fig}
\vspace{-10pt}
\end{figure}
An animated 3D blockout represents a shot using simple geometry, stand-in characters, and a camera. Its editable structure lets creators adjust camera motion, spatial layout, and character blocking for previsualization~\cite{hu2026previzwhiz}. Recent video generation methods use 3D layouts, scene geometry,
and human motion as structural
guidance~\cite{wang2025cinemaster,cao2025uni3c,tan2026unimoca, gen-1, alhaija2025cosmos, wang2023videocomposer}.
Rendering an editable blockout provides a practical way to
translate scene-level edits into such guidance. This workflow faces two related bottlenecks. For creators, manually reconstructing a blockout from a reference video requires time and 3D production expertise~\cite{huang2024blenderalchemy, gu2025blendergym}. For model training, producing blockouts that match target videos is costly, limiting the scale of paired data~\cite{wang2025cinemaster, cao2025uni3c}. As shown in \cref{fig:first_fig}, automatically reconstructing blockouts from existing videos can provide editable scenes for creators and aligned blockout--video pairs for training. This requires efficient, faithful, and scalable reconstruction.

However, two gaps remain in video-to-blockout reconstruction. First, existing video reconstruction benchmarks focus on indoor
scans of furnished spaces~\cite{tang2026bvb}, which differ
substantially from film shots with expressive camera motion,
staged character interactions, and precisely timed actions.
A dedicated benchmark is needed to assess reconstruction
fidelity across these cinematic dimensions. Second, although agents can inspect videos and operate Blender to build scenes~\cite{tang2026bvb, yin2026vision}, their judgments of geometric improvements can be unreliable~\cite{li2026lego}. Character animation adds further difficulty, requiring faithful poses, trajectories, and timing throughout the shot.

We introduce \bench, a benchmark for evaluating camera motion, spatial layout, and character blocking in video-to-blockout reconstruction. It contains 50 single shots from 50 films and 416 keyframes sampled at approximately 1.5-second intervals, including shot endpoints. Camera references, depth pseudo ground truth, and human annotations enable quantitative comparison between reconstructed Blender scenes and the source videos. We also report runtime and API cost to assess efficiency.

Recent advances in visual geometry
recovery~\cite{DA3,wang2025vggt} and world-grounded human
motion estimation~\cite{wang2024tram,shen2024world}
provide a foundation for grounding reconstruction in
measured observations. We propose \method to ground agentic construction in measured observations and explicit checks. \emph{Metric Observation} first recovers cameras, human motion, and geometric evidence in a shared metric coordinate system. \emph{Agentic Modeling} then uses this evidence and the source video to construct the environment from named, editable primitives. Recovered cameras and human poses remain fixed, with constrained adjustments for human scale and support alignment. \emph{Evidence-Based Verification} reports depth conflicts, missing structural coverage, and inconsistent human--environment relationships to guide scene refinement. On \bench, \method achieves an overall score of up to 76.1, improving on the corresponding vanilla baselines by 27.5–41.1 points across three agents. With DeepSeek-1017, \method achieves a score of 73.2 at an average API cost of only \$0.054 per shot, surpassing the strongest vanilla baseline (GPT-6.1 Sol, 48.6) while reducing total reconstruction time by 53\%.

Using \method, we construct \dataset, a dataset of 3,000 blockout--video pairs for public release. Each reconstructed scene is rendered with the recovered camera trajectory to produce a blockout video aligned frame by frame with the source, providing paired data for training blockout-conditioned video generation models. The reconstructed scenes also let creators adapt a reference shot in Blender by changing the viewpoint and revising character paths. Creators can preview these variants and render them as blockout videos to condition video generation.

Our contributions are:
\begin{itemize}
    \item \bench, a benchmark for quantitative evaluation of camera motion, spatial layout, and character blocking in video-to-blockout reconstruction.
    \item \method, a framework combining metric observation, agentic modeling, and evidence-based verification to reconstruct editable blockouts.
    \item \dataset, 3,000 blockout--video pairs for public release, with applications to previsualization editing and blockout-conditioned video generation.
\end{itemize}
\section{Benchmark}
\label{sec:benchmark}

\subsection{Task Formulation}
\label{sec:task}
Given a single film shot $V=\{I_t\}_{t=1}^{N}$ of 3--20\,s featuring people, where $N$ denotes the total number of frames, the task is to reconstruct an editable 3D blockout scene $\mathcal{B}$ and render it into a corresponding blockout video $\hat{V}$:
\begin{equation}
  \mathcal{B}=(\mathcal{C},\mathcal{G},\mathcal{H})=M(V),
  \qquad \hat{V}=\operatorname{Render}(\mathcal{B}),
  \label{eq:task}
\end{equation}
where $M$ denotes the video-to-blockout reconstruction process, and $\operatorname{Render}$ renders the reconstructed scene using the recovered camera trajectory to reproduce the original shot in a blockout representation.

The blockout scene consists of three components:
the shot camera $\mathcal{C}=\{(K_t,R_t,\mathbf{t}_t)\}_{t=1}^{N}$, specifying the camera intrinsics and extrinsics at each frame;
the greybox scene $\mathcal{G}$, represented by a set of named geometric primitives, such as boxes, cylinders, and planes; and the stand-in characters $\mathcal{H}=\{\mathbf{h}^{k}_t\}$, where each person $k$ is represented by a rigged mannequin with a pose defined at every frame $t$.

The rendered blockout video $\hat{V}$ is required to match the input video in resolution, frame rate, and frame count, establishing a frame-aligned video pair $(V,\hat{V})$. The underlying 3D scene must remain editable, allowing users to modify the camera, scene geometry, and character poses. 
\begin{figure}[t]
    \centering
    \includegraphics[width=\columnwidth]{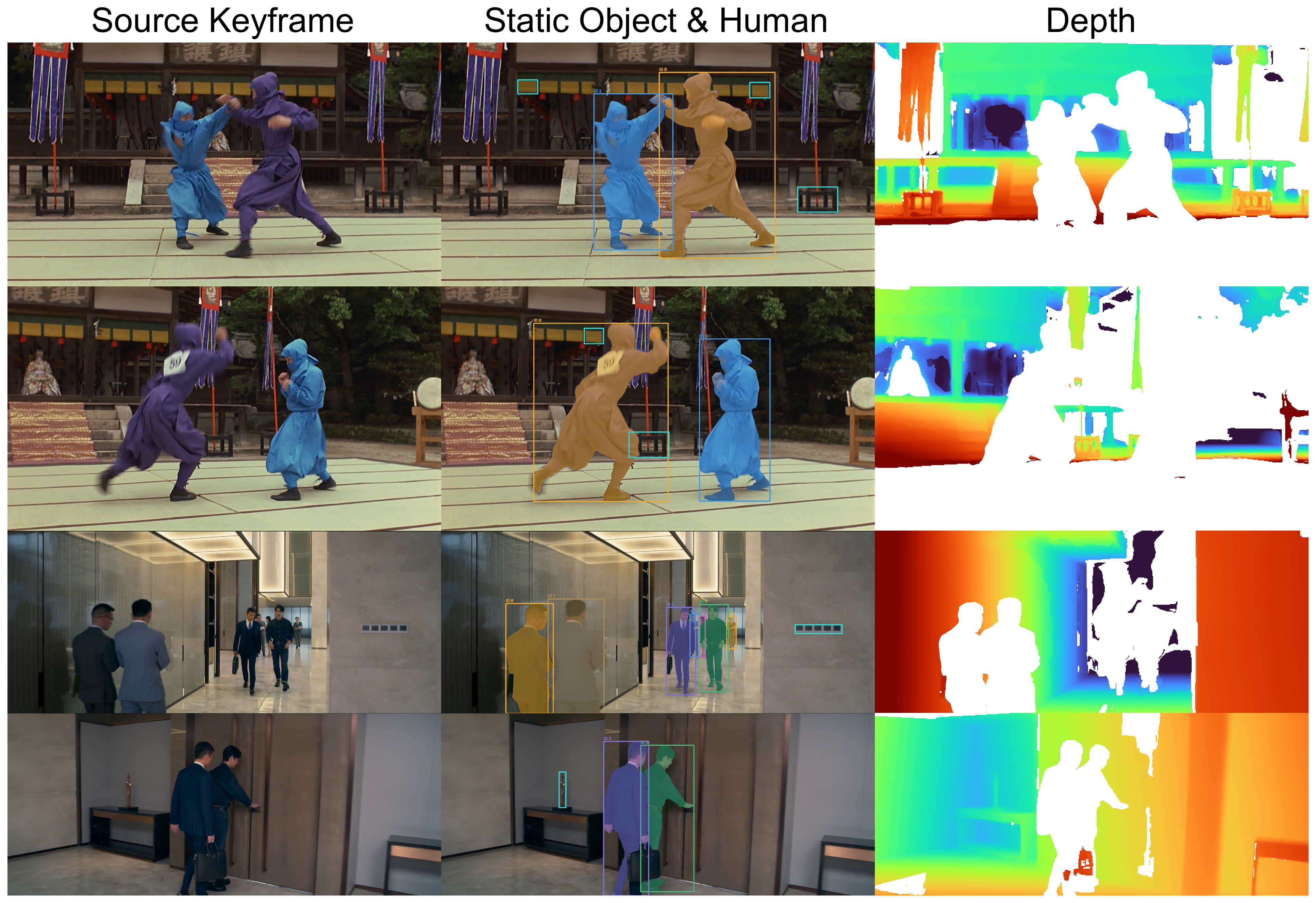}
    \caption{\textbf{Annotation examples from \bench.}
    Columns show source keyframes, annotated static-object boxes and person identities and masks, and depth pseudo ground truth. Person colors indicate identities within each shot. White regions in the depth maps are excluded from depth evaluation.}
    \label{fig:annotations}
\vspace{-4pt}
\end{figure}

\subsection{Data Collection and Annotation}
\label{sec:data_annotation}
\bench contains 50 single shots from 50 films, covering
diverse camera movements, human actions, spatial relationships,
and multi-person configurations.
We sample 416 keyframes at approximately 1.5-second intervals,
including shot endpoints.
Reference cameras are obtained through cross-validation of
multiple reconstruction methods, with manually annotated
static-object tracks supporting ORE evaluation.
Depth pseudo ground truth is constructed by combining
DA3~\cite{DA3}, $\pi^3$~\cite{wang2026pi}, and VGGT~\cite{wang2025vggt}
predictions with geometric consistency filtering.
Person identities and masks are initialized by
SAM~3~\cite{carion2026sam} and manually corrected.
\cref{fig:annotations} shows annotation examples.
Shot selection, annotation procedures, and filtering details
are provided in the supplementary material.

\subsection{Evaluation Protocol}
\label{sec:bench_metrics}
We evaluate camera motion, spatial layout, and character blocking directly from the submitted \texttt{.blend} files. For each shot, we render instance and depth maps using the camera in the reconstructed scene at 1.5\,s intervals, yielding 416 keyframes in total. Metrics are computed for each shot and uniformly averaged across shots.
\paragraph{Camera Motion.}
We evaluate camera motion using Object Reprojection Error (ORE)$\downarrow$~\cite{ore} and Relative Rotation and Translation Accuracy at $15^\circ$ (RRA@15 and RTA@15)$\uparrow$~\cite{posediffusion,wang2025vggt}. ORE measures cross-frame reprojection consistency of static objects without reference camera trajectories and is evaluated on all 50 shots. RRA@15 and RTA@15 measure the fractions of frame pairs whose relative rotation and translation direction, respectively, fall within $15^\circ$ of the reference camera trajectories.

\paragraph{Spatial Layout.}
We evaluate relative depth ordering using ordinal accuracy~\cite{DIW}. For each keyframe, we sample 40k pixel pairs and measure the fraction whose predicted depths preserve the ground-truth ordering. Ties and missing reconstructed geometry count as errors.

\paragraph{Character Blocking.}
We render character silhouettes with scene occlusion. We report HOTA$\uparrow$~\cite{luiten2021hota}, including its detection (DetA) and association (AssA) components, using keyframes as frames and box IoU as the similarity measure. HOTA captures missing, spurious, and identity-switched characters. We further evaluate character silhouettes using Mask IoU$\uparrow$~\cite{davis}, which measures the overlap between rendered and ground-truth masks. Within each shot, annotated persons are matched to proxies via Hungarian matching~\cite{hungarian} based on accumulated bounding-box IoU. We average Mask IoU over all person–keyframe instances, assigning zero to unmatched instances.

\paragraph{Overall Score.}
The overall score is the mean of the three axis scores multiplied by 100. The camera score is the average of $1-\min(1,\mathrm{ORE}/\mathrm{ORE}_{\mathrm{static}})$, RRA@15, and RTA@15 over the available metrics for each shot, where $\mathrm{ORE}_{\mathrm{static}}=0.168$ approximates the ORE of a static camera. The depth score is ordinal accuracy, while the character score is the average of HOTA and Mask IoU. The overall score is the mean of the three axis scores.

\paragraph{Efficiency.}
We report mean API cost and wall-clock runtime per shot, from reconstruction start to Blender scene completion, excluding final video rendering. 

\section{Method}
\label{sec:method}
\begin{figure*}[t]
    \centering
    \includegraphics[width=\textwidth]{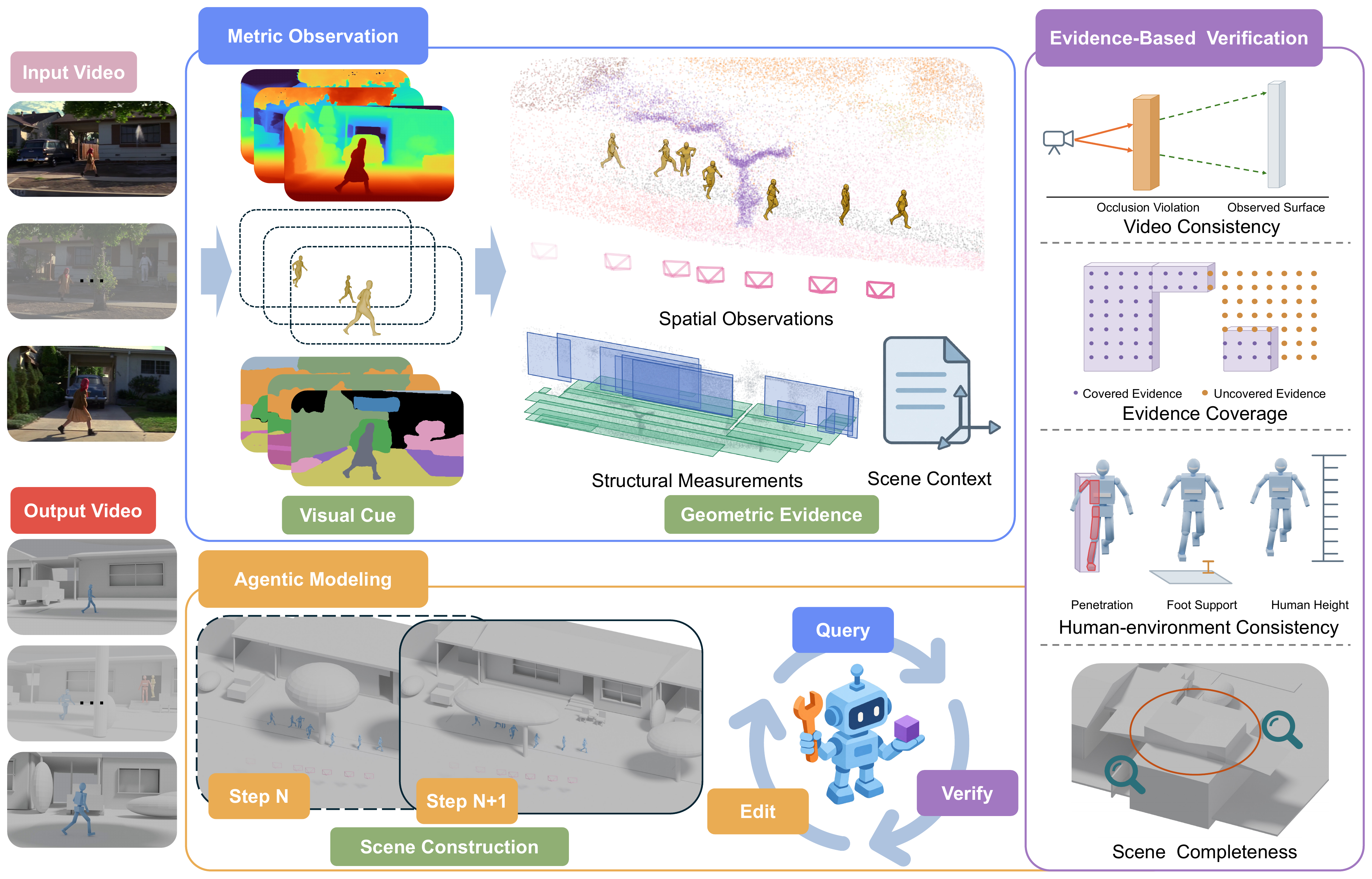}
    \caption{\textbf{Overview of \method.}
        Metric Observation recovers cameras and human motion in a shared metric coordinate system and extracts geometric evidence. Agentic Modeling constructs an editable environment through iterative queries and scene edits. Evidence-Based Verification checks video consistency, evidence coverage, human--environment consistency, and scene completeness, providing feedback for refinement. The reconstructed scene is rendered into a blockout video aligned with the input.}
\label{fig:overview}
\end{figure*}

To reduce the capability requirements and costs of agentic scene construction, we propose a structured framework for 3D blockout reconstruction, as shown in~\cref{fig:overview}. The framework first recovers the camera trajectory and metric character geometry in a shared 3D coordinate system. An LLM-based agent then iteratively constructs the environment, editing the scene and verifying its geometry against the input video to produce a coherent, editable 3D scene.

\begin{figure*}[t]
    \centering
    \includegraphics[width=\textwidth]{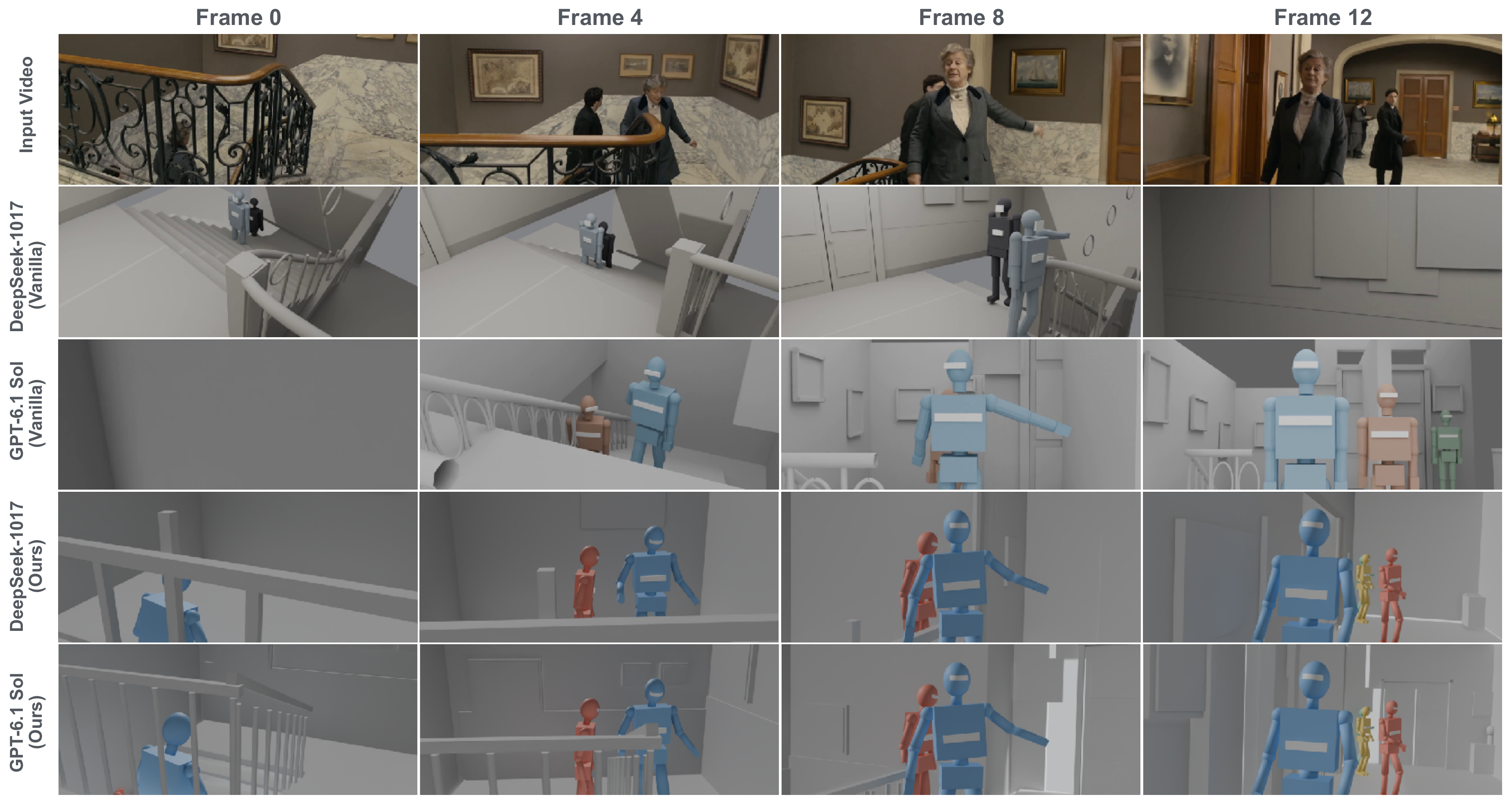}
    \caption{\textbf{Qualitative comparison on a film shot.}
    Rows show the input video, Vanilla Agent reconstructions with DeepSeek-1017 and GPT-6.1 Sol, and \method reconstructions with the same agents. Columns show corresponding frames over time.}
\label{fig:qualitative}
\end{figure*}
\subsection{Metric Observation}
\label{sec:observation}

Given an input video, we recover video-aligned camera and human representations, denoted by $\mathcal{C}$ and $\mathcal{H}$, respectively, in a metric coordinate system. We also extract geometric evidence $\mathcal{D}$ to support subsequent scene construction and agentic reasoning.

\paragraph{Camera and human recovery.}
We first apply DA3~\cite{DA3} to uniformly sampled keyframes to jointly estimate depth maps, camera intrinsics, and camera poses in a shared coordinate system. We interpolate the camera poses and intrinsics to obtain the full video trajectory. Meanwhile, SAM~3~\cite{carion2026sam} segments and tracks people throughout the video, providing frame-wise person masks and bounding boxes. Given these masks, bounding boxes, and the corresponding DA3 camera intrinsics, SAM~3D Body~\cite{yang2026sam} estimates metric 3D human poses in the camera coordinate system. This produces video-aligned camera and human representations $\mathcal{C}$ and $\mathcal{H}$.

\paragraph{Metric scene alignment.}
We first align the scale of DA3 depth with the metric human estimates from SAM~3D Body~\cite{muller2025reconstructing}. For each person $k$ in keyframe $j$, we estimate the corresponding visible-surface depths $z^{\mathrm{body}}_{k,j}$ and $z^{\mathrm{DA3}}_{k,j}$ from the projected human meshes and DA3 depth maps, respectively~\cite{muller2025reconstructing, yalandur2025physic}. The global scale factor is computed as
\begin{equation}
    s = \operatorname*{median}_{k,j}
    \left(
        z^{\mathrm{body}}_{k,j}
        \big/
        z^{\mathrm{DA3}}_{k,j}
    \right).
    \label{eq:metric-scale}
\end{equation}
We apply $s$ to the DA3 depth maps and camera translations to obtain a metric scene representation.

Gravity alignment provides a consistent reference for
world-space human motion~\cite{shen2024world}. We estimate the scene up direction $\mathbf{u}^{\star}$ by combining geometric and orientation cues:
\begin{equation}
\begin{aligned}
\mathbf{u}^{\star}
&= \arg\min_{\|\mathbf{u}\|=1}
\sum_{a\in\{\mathrm{g},\mathrm{ceil}\}}
\psi\left(\frac{\angle(\mathbf{u},\mathbf{n}_a)}{\gamma_a}\right)
+ \frac{\angle(\mathbf{u},\mathbf{v}_{\mathrm{cam}})^2}{\gamma_{\mathrm{cam}}^2}
\\
&\quad + \sum_w \psi\left(\frac{\arcsin|\mathbf{u}^{\top}\mathbf{n}_w|}{\gamma_w}\right)
+ \psi\left(\frac{\angle(\mathbf{u},\mathbf{v}_{\mathrm{body}})}{\gamma_{\mathrm{body}}}\right).
\end{aligned}
\label{eq:gravity}
\end{equation}

Here, $\mathbf{n}_{\mathrm{g}}$ and $\mathbf{n}_{\mathrm{ceil}}$ denote ground and ceiling normals, both oriented upward, $\mathbf{n}_w$ denotes vertical-surface normals, $\mathbf{v}_{\mathrm{body}}$ is the human upright direction, and $\mathbf{v}_{\mathrm{cam}}$ is the camera up direction. The $\gamma$ terms are scale parameters, and the robust penalty $\psi$ reduces the influence of unreliable observations.

Finally, we construct a unified scene coordinate system with $\mathbf{u}^{\star}$ as its positive $z$-axis. We fit a ground plane from the static scene point cloud and place the origin at the projection of the first camera center onto this plane. The horizontal axes are initialized from the first camera orientation and rotated around the $z$-axis to align with the dominant Manhattan directions~\cite{coughlan2003manhattan} estimated from static scene normals.

\paragraph{Human placement.}
After scene alignment, we place each person using DA3 depth.
We estimate the neck depth $\bar{d}_{k,t}$ from the aligned
depth within the person mask and compute a per-frame position factor~\cite{smap}:
\begin{equation}
    f_{k,t} = \bar{d}_{k,t} / z^{\mathrm{b}}_{k,t},
    \quad
    m_k = \operatorname{median}_{t} f_{k,t},
    \label{eq:human-scale}
\end{equation}
where $z^{\mathrm{b}}_{k,t}$ is the neck depth predicted by SAM~3D Body. Let $\mathbf{J}_{k,t}$ denote its predicted 3D body keypoints and $\mathbf{J}^{\mathrm{neck}}_{k,t}$ the neck keypoint. We place the human in the camera coordinate system as
\begin{equation}
    \mathbf{P}_{k,t}
    =
    f_{k,t}\mathbf{J}^{\mathrm{neck}}_{k,t}
    +
    m_k
    \left(
        \mathbf{J}_{k,t}
        -
        \mathbf{J}^{\mathrm{neck}}_{k,t}
    \right).
    \label{eq:human-placement}
\end{equation}
Here, $f_{k,t}$ adjusts the neck position along its camera ray to match the DA3 depth, while the temporally constant $m_k$ scales the body around the neck, preserving a consistent human height throughout the video. The resulting poses are then transformed into the scene coordinate system.

\paragraph{Geometric evidence extraction.}
We back-project the aligned DA3 depth maps into the scene coordinate system to construct a static point cloud $\mathcal{Q}$, excluding dynamic human regions and unreliable depth observations. We then assign semantic labels to the retained points using Mask2Former~\cite{mask2former} and aggregate them spatially to extract wall segments $\mathcal{W}$, support layers $\mathcal{L}$, and ceiling regions $\mathcal{U}$. Together with camera and human information, these observations form the geometric evidence set
\begin{equation}
    \mathcal{D}
    =
    \left\{
        \mathcal{Q},
        \mathcal{W},
        \mathcal{L},
        \mathcal{U},
        \mathcal{C},
        \mathcal{H}
    \right\}.
    \label{eq:geometric-evidence}
\end{equation}
This evidence provides geometric and spatial references for subsequent agentic scene construction and verification.

\subsection{Agentic Modeling}
\label{sec:agent}

\paragraph{Scene initialization.}
We initialize the scene with the recovered cameras and human representations, leaving the environment geometry empty:
\begin{equation}
    \mathcal{B}_0=(\mathcal{C},\varnothing,\mathcal{H}).
    \label{eq:initial-scene}
\end{equation}

\paragraph{Scene construction.}
Given the source video and geometric evidence $\mathcal{D}$, the agent constructs the environment using editable parametric primitives:
\begin{equation}
    \sigma=(\kappa,\mathbf{c},\mathbf{e},\theta,\tau,\nu),
    \label{eq:primitive}
\end{equation}
where $\kappa$, $\mathbf{c}$, $\mathbf{e}$, and $\theta$ denote the primitive type, center, dimensions, and orientation, while $\tau$ and $\nu$ specify its semantic class and name. Complex structures combine multiple primitives~\cite{tulsiani2017learning,ye2025primitiveanything}. At each iteration, Blender executes an agent-generated edit program $e_n$ to update the scene:
\begin{equation}
    \mathcal{B}_n
    =
    \operatorname{Apply}(\mathcal{B}_{n-1},e_n).
    \label{eq:scene-update}
\end{equation}
Camera trajectories and human poses remain fixed, while each human track has a single scale parameter $\lambda_k$ that jointly controls its physical size and camera distance while preserving its image projection. The scale remains constant throughout the video.

\paragraph{Human-environment alignment.}
The video constrains human poses and image projections but does not uniquely determine their metric scale~\cite{PHOSA,MOVER}. We parameterize each human track with a single scale factor $\lambda_k$, which scales the entire trajectory about the corresponding camera centers:
\begin{equation}
    \mathbf{P}'_{k,t}
    =
    \mathbf{o}_t
    +
    \lambda_k
    \left(
        \mathbf{P}_{k,t}-\mathbf{o}_t
    \right).
    \label{eq:human-rescaling}
\end{equation}
Here, $\mathbf{o}_t$ is the camera center at frame $t$.
This transformation preserves image projection while scaling the person's height and camera distance~\cite{Hartley2004}. A single $\lambda_k$ is shared across all frames.

The agent determines $\lambda_k$ through three operations: (1) support-based placement, which aligns the feet with a specified scene primitive; (2) height-based placement, which sets $\lambda_k=H^\star/H_k$ given an estimated physical height $H^\star$; and (3) direct assignment. Since the same scale factor governs both human height and support alignment, the resulting height provides an additional check on the reconstructed geometry. Inconsistent constraints prompt the agent to reconsider the scene geometry or human interpretation. The agent can also remove entire tracks or invalid temporal segments identified as non-human objects or tracking failures, without modifying the remaining poses.

\paragraph{Closed-loop modeling.}
Following iterative code--render--inspect workflows for
agentic graphics~\cite{yin2026vision, huang2024blenderalchemy},
we organize scene construction into a query--edit--verify loop. The agent inspects the scene, queries evidence, executes edits, and analyzes diagnostic renderings until completion or exhaustion of the time budget. At each iteration, the agent selects an operation according to
\begin{equation}
    e_n
    \sim
    \pi_{\mathrm{LLM}}
    \left(
        \cdot\mid
        \mathcal{T},\mathcal{D},\mathcal{R}_{n-1}
    \right),
    \label{eq:agent-policy}
\end{equation}
where $\mathcal{T}$ denotes the task description and tool documentation, and $\mathcal{R}_{n-1}$ records previous queries, edits, and verification feedback.

\subsection{Evidence-Based Verification}
\label{sec:verify}
Motivated by the limitations of agents' geometric
self-assessment~\cite{li2026lego}, we provide explicit
checks against measured scene evidence. After each scene edit, verification tools evaluate the reconstruction against the input video and geometric evidence, reporting inconsistencies through diagnostic feedback. The agent resolves these issues through further edits or explicit justifications. We consider four complementary aspects.

\paragraph{Video consistency.}
We check whether reconstructed primitives incorrectly occlude surfaces observed in the input video. 
For each primitive $\sigma$, we compare its projected depth with
the observed depth across keyframes and quantify potential
occlusion violations by measuring the proportion of overlapping
pixels where the primitive lies significantly in front of
the observed surface:
\begin{equation}
    \phi(\sigma)
    =
    \frac{
    \sum_j
    \left|
    \left\{
    \mathbf{x}\in\Omega_j\cap\Omega_j^\sigma:
d_\sigma(\mathbf{x})<d_j(\mathbf{x})-\delta(d_j(\mathbf{x}))
    \right\}
    \right|
    }{
    \sum_j|\Omega_j\cap\Omega_j^\sigma|
    },
    \label{eq:occlusion-check}
\end{equation}
where $\Omega_j$ and $\Omega_j^\sigma$ denote the valid observed pixels and projected primitive region, respectively; $d_j$ and $d_\sigma$ are their depths, and $\delta$ is a depth-dependent tolerance. Primitives with substantial violations are flagged for inspection.

\paragraph{Evidence coverage.}
We assess whether scene primitives spatially cover the geometric measurements in $\mathcal{D}$, including wall segments $\mathcal{W}$, support layers $\mathcal{L}$, and ceiling regions $\mathcal{U}$, subject to geometric proximity constraints. Uncovered or misaligned measurements are reported to the agent for reconstruction or justification based on the source video.

\paragraph{Human--environment consistency.}
We check body penetration, foot support, and estimated human height to verify the spatial relationships between reconstructed people and scene geometry~\cite{PROX,MOVER, ppr}. The agent uses the reported violations to revise scene geometry, adjust the human scale $\lambda_k$, or justify exceptional poses such as jumping or riding. Removed tracks and invalid temporal segments also require explicit justifications.

\paragraph{Scene completeness.}
Using orbit views, the agent inspects global scene structure, including spatial enclosure, openings, and floating, duplicated, or disconnected geometry. This complements local geometric checks with an assessment of overall structural coherence. 
\paragraph{Submission and output.}
The submission tool checks whether reported issues have been
resolved or justified. At the time limit, the latest successful
scene is retained with unresolved issues recorded.
The final scene is rendered using the recovered cameras:
\begin{equation}
    \hat{I}_t =
    \operatorname{Render}
    \left(\mathcal{B}_{\mathrm{final}};K_t,R_t,\mathbf{t}_t\right),
    \hat{V}=\{\hat{I}_t\}_{t=1}^{N}.
    \label{eq:final-render}
\end{equation}
We deliver the editable scene and its blockout video at the
source resolution, frame rate, and frame count, rendering
characters only when visible in the source.

\begin{figure*}[!t]
    \centering
    \includegraphics[width=\textwidth]{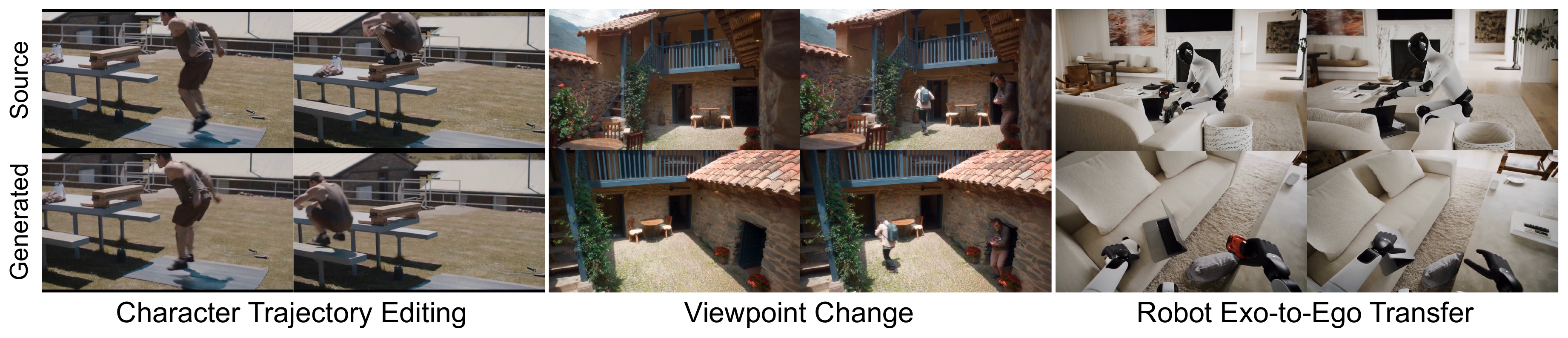}
    \caption{\textbf{Video generation guided by edited blockouts.}
    From left to right: character trajectory editing, viewpoint change, and robot exocentric-to-egocentric view transfer. The top row shows source frames, and the bottom row shows generated results conditioned on renders of the edited blockouts.}
    \label{fig:applications}
\end{figure*}

\begin{table*}[t]
    \centering
    \small
    \tabcolsep=0.45em
    \resizebox{\textwidth}{!}{
    \begin{tabular}{@{}l cccc c ccc c@{}}
        \toprule
        \multirow{2}[2]{*}{\textbf{Method}}
        & \multicolumn{4}{c}{\textbf{Camera Motion}}
        & \textbf{Spatial Layout}
        & \multicolumn{3}{c}{\textbf{Character Blocking}}
        & \multirow{2}[2]{*}{\textbf{Total}$\uparrow$} \\
        \cmidrule(lr){2-5}
        \cmidrule(lr){6-6}
        \cmidrule(lr){7-9}
        & ORE$\downarrow$
        & RRA@15$\uparrow$
        & RTA@15$\uparrow$
        & Score$\uparrow$
        & OrdAcc$\uparrow$
        & HOTA$\uparrow$
        & MaskIoU$\uparrow$
        & Score$\uparrow$
        & \\
        \midrule

        No Reconstruction
        & 0.168 & 0.000 & 0.000 & 0.000
        & 0.000
        & 0.000 & 0.000 & 0.000
        & 0.0 \\

        \midrule
        \multicolumn{10}{@{}l}{\textit{Vanilla}} \\

        GPT-6.1 Sol
        & \secondbest{0.060}
        & \secondbest{0.571}
        & 0.215
        & \secondbest{0.477}
        & 0.609
        & 0.421 & 0.324 & 0.373
        & 48.6 \\

        Claude Opus 5.5
        & 0.077 & 0.542
        & \secondbest{0.247}
        & 0.444
        & 0.585
        & 0.264 & 0.175 & 0.220
        & 41.6 \\

        DeepSeek-1017
        & 0.123 & 0.430 & 0.184 & 0.293
        & 0.596
        & 0.260 & 0.191 & 0.226
        & 37.2 \\

        Claude Sonnet 5.5
        & 0.135 & 0.346 & 0.209 & 0.250
        & 0.534
        & 0.195 & 0.131 & 0.163
        & 31.6 \\

        Kimi K3.1 Pro
        & 0.156 & 0.208 & 0.185 & 0.155
        & 0.511
        & 0.215 & 0.152 & 0.183
        & 28.3 \\

        \midrule
        \multicolumn{10}{@{}l}{\textbf{\method (Ours)}} \\

        GPT-6.1 Sol
        & \best{0.005}
        & \best{1.000}
        & \best{0.844}
        & \best{0.938}
        & \best{0.781}
        & \best{0.627}
        & \best{0.499}
        & \best{0.563}
        & \best{76.1} \\

        DeepSeek-1017
        & \best{0.005}
        & \best{1.000}
        & \best{0.844}
        & \best{0.938}
        & \secondbest{0.717}
        & 0.604 & 0.478 & 0.541
        & \secondbest{73.2} \\

        Claude Sonnet 5.5
        & \best{0.005}
        & \best{1.000}
        & \best{0.844}
        & \best{0.938}
        & 0.695
        & \secondbest{0.612}
        & \secondbest{0.485}
        & \secondbest{0.548}
        & 72.7 \\

        \bottomrule
    \end{tabular}
    }
    \caption{\textbf{Reconstruction quality on \bench.}
    Results are averaged over 50 film shots and evaluate camera motion, spatial layout, and character blocking. Total is the mean of the three axis scores, expressed on a 0--100 scale. All \method runs share the cameras recovered by Metric Observation. Best and second-best results among reconstruction methods are highlighted.}
    \label{tab:main_results}
\end{table*}

\section{Experiments}
\label{sec:experiments}

\subsection{Experimental Setup}
\label{sec:exp_setup}

We evaluate all 50 shots in \bench following the protocol described in \cref{sec:bench_metrics}.
Each agent is allocated a 15-minute target runtime, with a hard limit of 30 minutes. The most recent successfully executed scene is used for evaluation.
All agents operate with high reasoning effort.

\emph{No Reconstruction} renders an empty scene using a static camera.
\emph{Vanilla Agent} allows an LLM agent to construct the scene in Blender using only the input video and a shared task description, without access to the observation, evidence, and verification tools provided by \method.

We evaluate five LLM agents in the Vanilla setting: GPT-6.1 Sol~\cite{openai2026gpt61sol}, Claude Opus 5.5~\cite{anthropic2026claudeopus55}, DeepSeek-V4.1-Flash-1017~\cite{deepseek2026v41flash}, Claude Sonnet 5.5~\cite{anthropic2026claudesonnet55}, and Kimi K3.1 Pro~\cite{moonshot2026kimik31pro}. We further evaluate GPT-6.1 Sol, DeepSeek-V4.1-Flash-1017, and Claude Sonnet 5.5 with \method.

\subsection{Main Results}
\label{sec:main_results}

\paragraph{Reconstruction Quality.}
As shown in \cref{tab:main_results}, \method consistently improves all three agents, yielding gains of 27.5--41.1 points in overall score. All three outperform the strongest Vanilla result (48.6). The largest gains are observed in camera motion, with score improvements of 0.461--0.688, followed by character blocking (0.190--0.385) and spatial layout (0.121--0.172). Moreover, \method benefits weaker agents more, reducing the performance gap among the three agents from 17.0 to 3.4 points.

\paragraph{Efficiency.}
\method achieves substantial quality improvements with comparable API costs (\cref{tab:efficiency}). The per-shot API cost changes by at most \$0.06 across the three agents. Meanwhile, agent execution time decreases by 42\% for GPT-6.1 Sol and 55\% for DeepSeek-1017, partially offsetting the additional 1m28s required by Metric Observation. Notably, DeepSeek-1017 with \method surpasses GPT-6.1 Sol with Vanilla by 24.6 points in overall score, while reducing API cost by 86\% and total runtime by 53\%.

\begin{table}[t]
    \centering
    \small
    \tabcolsep=0.4em
    \resizebox{\linewidth}{!}{
    \begin{tabular}{@{}l cccc@{}}
        \toprule
        \textbf{Method}
        & Agent Time $\downarrow$
        & \# Primitives
        & Cost $\downarrow$
        & Total$\uparrow$ \\
        \midrule
        \multicolumn{5}{@{}l}{\textit{Vanilla}} \\
        GPT-6.1 Sol
        & 22m41s & 273.1 & \$0.385 & 48.6 \\
        Claude Opus 5.5
        & 15m50s & 36.4 & \$0.644 & 41.6 \\
        DeepSeek-1017
        & 20m25s & 182.3 & \$\secondbest{0.074} & 37.2 \\
        Claude Sonnet 5.5
        & \best{3m46s} & 67.7 & \$0.341 & 31.6 \\
        Kimi K3.1 Pro
        & 24m53s & 271.2 & \$0.368 & 28.3 \\
        \midrule
        \multicolumn{5}{@{}l}{\textbf{\method (Ours)} \hfill \textit{+ Metric Observation: 1m28s}} \\
        GPT-6.1 Sol
        & 13m11s & 115.5 & \$0.445 & \best{76.1} \\
        DeepSeek-1017
        & 9m17s & 68.3 & \best{\$0.054} & \secondbest{73.2} \\
        Claude Sonnet 5.5
        & \secondbest{4m36s} & 20.9 & \$0.363 & 72.7 \\
        \bottomrule
    \end{tabular}
    }
    \caption{\textbf{Reconstruction efficiency and scene complexity on \bench.}
    \# Primitives counts scene primitives.
    \method requires an additional 1m28s for Metric Observation.}
    \label{tab:efficiency}
\end{table}

\subsection{Ablation Studies}
\label{sec:ablation}

\paragraph{Components.}
We ablate each component using DeepSeek-1017 (\cref{tab:ablation}). All variants retain the recovered cameras, which establish the metric frame shared by depth, geometric evidence, and human placement. Removing camera recovery would disable these components, reducing the method to Vanilla and preventing their individual contributions from being isolated. We fix the cameras and focus on spatial layout and character blocking.

\emph{w/o Geometric Evidence} removes the point cloud, depth map, and extracted structural elements (walls, supports, and ceilings), requiring the agent to reconstruct the environment from scratch using only the input video. Pre-placed humans are retained.
\emph{w/o Human Recovery} removes the pre-placed human proxies, requiring the agent to create and pose them independently.
\emph{w/o Verification} removes the verification tools and submission checks.

Removing geometric evidence reduces OrdAcc by 0.129, while leaving character blocking largely unaffected. Removing human recovery causes the largest overall score drop (12.0 points), with HOTA and Mask IoU decreasing by 0.363 and 0.300, respectively. Verification provides smaller but consistent gains across both axes, improving the overall score by 2.2 points. Together, these results demonstrate the complementary contributions of the three components.

\paragraph{Time Budget.}
\method achieves an overall score of 72.8 with a 5-minute target, close to the best score of 73.2 obtained with the default 15-minute target (\cref{tab:time_budget}). Extending the target to 30 minutes brings no further improvement. Moreover, even with a 30-minute target and a 40-minute hard limit, the agent finishes in 10m25s on average, indicating that \method achieves strong reconstruction quality without requiring prolonged agent interaction.

Additional experiments and analyses, including a user study, are provided in the appendix.
\begin{table}[t]
    \centering
    \small
    \tabcolsep=0.4em
    \resizebox{\linewidth}{!}{
    \begin{tabular}{l c cc c}
        \toprule
        \multirow{2}[2]{*}{\textbf{Variant}}
        & \textbf{Layout}
        & \multicolumn{2}{c}{\textbf{Blocking}}
        & \multirow{2}[2]{*}{\textbf{Total}$\uparrow$} \\
        \cmidrule(lr){2-2}
        \cmidrule(lr){3-4}
        & OrdAcc$\uparrow$
        & HOTA$\uparrow$
        & MaskIoU$\uparrow$
        & \\
        \midrule
        \rowcolor{refcolor}
        \method (full)
        & 0.717 & 0.604 & 0.478 & 73.2 \\

        w/o Geometric Evidence
        & \cellcolor[rgb]{1.000,0.500,0.500}{$-$0.129}
        & \cellcolor[rgb]{1.000,0.982,0.982}{$-$0.013}
        & \cellcolor[rgb]{1.000,0.995,0.995}{$-$0.003}
        & \cellcolor[rgb]{1.000,0.808,0.808}{$-$4.6} \\

        w/o Human Recovery
        & \cellcolor[rgb]{1.000,0.884,0.884}{$-$0.030}
        & \cellcolor[rgb]{1.000,0.500,0.500}{$-$0.363}
        & \cellcolor[rgb]{1.000,0.500,0.500}{$-$0.300}
        & \cellcolor[rgb]{1.000,0.500,0.500}{$-$12.0} \\

        w/o Verification
        & \cellcolor[rgb]{1.000,0.833,0.833}{$-$0.043}
        & \cellcolor[rgb]{1.000,0.960,0.960}{$-$0.029}
        & \cellcolor[rgb]{1.000,0.973,0.973}{$-$0.016}
        & \cellcolor[rgb]{1.000,0.908,0.908}{$-$2.2} \\
        \bottomrule
    \end{tabular}
    }
    \caption{\textbf{Component ablation with DeepSeek-1017.}
    Changes are relative to the full \method (gray).
    Darker red indicates larger drops.
    All variants share a camera score of 0.938, omitted here.}
    \label{tab:ablation}
\end{table}

\begin{table}[t]
    \centering
    \small
    \tabcolsep=0.45em
    \begin{tabular}{cc ccc c}
        \toprule
        \multicolumn{2}{c}{\textbf{Budget (min)}}
        & \multicolumn{3}{c}{\textbf{Score}$\uparrow$}
        & \multirow{2}[2]{*}{\textbf{Agent Time}} \\
        \cmidrule(lr){1-2}
        \cmidrule(lr){3-5}
        Target & Limit
        & Layout & Blocking & Total
        & \\
        \midrule
        5  & 10 & \bestab{0.720} & 0.525 & 72.8 & 6m42s \\
        \rowcolor{refcolor}
        15 & 30 & 0.717 & \bestab{0.541} & \bestab{73.2} & 9m17s \\
        30 & 40 & 0.718 & 0.538 & 73.1 & 10m25s \\
        \bottomrule
    \end{tabular}
    \caption{\textbf{Agent time budget.}
    Layout and Blocking are the spatial layout and character blocking scores.
    Gray: default setting. All runs use DeepSeek-V4.1-Flash-1017.}
    \label{tab:time_budget}
\end{table}

\subsection{Qualitative Results}
\label{sec:qualitative}
\cref{fig:qualitative} shows a shot in which the camera follows people up a staircase and along a corridor, with four people appearing over the course of the shot.
With Vanilla Agent, the reconstructed cameras deviate from the input views, \eg, ending on a blank wall or starting behind geometry.
With \method, both agents follow the source camera, and the stairs, railing, and people stay aligned with the input across frames. In particular, \method closely reproduces the human motions in the input video. Additional qualitative results are provided in the appendix.

\section{Applications}
\label{sec:applications}

\paragraph{\dataset.}
We construct \dataset, comprising 3,000 blockout--video
pairs from 410 films, with at most 12 source clips per film.
Starting from approximately 130K raw film clips,
initial filtering and visual screening yield 12,050
candidates, from which we select the final 3,000 distinct clips.
The selected clips total 6.19 hours, with durations
ranging from 3 to 15 seconds (mean: 7.43 seconds).
Camera motion and human actions occur in 38.6\% and
38.5\% of clips, respectively, with 64.8\% containing
at least one of the two. Single-person and two-person
shots account for 31.4\% and 34.0\%, respectively;
the remaining clips feature groups of three to six people
(24.0\%) or crowds (10.6\%). Close, medium, and long shots
account for 19.8\%, 46.7\%, and 33.5\%, respectively.

We reconstruct the clips using \method with DeepSeek-1017,
GPT-6.1 Sol, and Claude Sonnet 5.5. Each source clip is
paired with a blockout video rendered at the same resolution,
frame rate, and frame count, and accompanied by an editable
Blender scene. Further details, including generation success
rates and the manual screening procedure, are reported in
the appendix. We will publicly release \dataset to support
the training of blockout-conditioned video generation models.

\paragraph{Blockout Editing and Video Generation.}
As shown in \cref{fig:applications}, reconstructed blockouts
support character trajectory editing, viewpoint changes,
and even robot exocentric-to-egocentric view transfer.
For a trajectory edit, creators modify a character's
path within the same scene. For a viewpoint change, creators adjust the camera while retaining the reconstructed environment and character
animation. We render the edited scenes into blockout videos that jointly represent camera motion, spatial layout, and character blocking, and use these videos to condition Seedance~2.5~\cite{bytedance2026seedance25} for video generation. Further details on blockout conditioning and comparisons with video generation without blockout conditioning are provided in the appendix.
\section{Related work}
\label{sec:related_work}

\paragraph{Programmatic Scene Construction.}
Programmatic scene construction provides a natural interface
for creating and editing structured 3D scenes.
LayoutGPT~\cite{feng2023layoutgpt} and
Holodeck~\cite{yang2024holodeck} translate language into
spatial layouts and asset arrangements, while
3D-GPT~\cite{sun20253d} connects language models with
procedural modeling tools.
SceneCraft~\cite{scenecraft} and
BlenderLLM~\cite{du2024blenderllm} extend this direction
to the generation of executable Blender programs.
For reconstruction from visual references, a central
challenge is determining whether program edits improve
the resulting scene.
BlenderAlchemy~\cite{huang2024blenderalchemy} and
VIGA~\cite{yin2026vision} use iterative execution and
visual feedback to refine graphics programs.
LEGO-Anything~\cite{li2026lego} studies image-to-code
scene reconstruction and highlights the difficulty
of judging geometric improvements.
Our framework grounds this iterative construction process
in geometric evidence recovered from video, using queryable
measurements and explicit checks to guide scene edits.

\paragraph{Benchmarks for Cinematic Understanding and Reconstruction.}
MovieNet~\cite{huang2020movienet},
CameraBench~\cite{lin2026towards}, and
ShotBench~\cite{liu2026shotbench} study movie understanding
and cinematographic attributes.
E.T.~\cite{ET} extracts camera and character
trajectories from movies for character-aware camera generation.
For programmatic graphics,
BlenderGym~\cite{gu2025blendergym} evaluates graphics editing,
while BVB~\cite{tang2026bvb} evaluates reconstruction of
indoor scan videos through question answering and perceptual
similarity.
In concurrent work CineRecon~\cite{wang2026cinerecon} benchmarks editable
dynamic scene reconstruction from film, real-world, and
game videos, using vision-language judgments of camera,
static, and dynamic consistency.
Our benchmark focuses on single film shots featuring people
and measures camera motion, spatial layout, and character
blocking through geometric and tracking metrics computed
from the delivered 3D scenes against reference annotations.

\paragraph{Controllable Video Generation and Previsualization.}
Controllable video generation introduces structural inputs
to complement textual descriptions.
CameraCtrl~\cite{he2024cameractrl} enables camera motion control, while MotionCtrl~\cite{wang2024motionctrl} supports control over both camera and object motion.
MotionCanvas~\cite{xing2025motioncanvas} supports cinematic
shot design through joint specification of camera and
object movements.
CineMaster~\cite{wang2025cinemaster} uses 3D object layouts,
rendered depth, and camera trajectories as generation
conditions.
Uni3C~\cite{cao2025uni3c} combines scene point clouds and
SMPL-X characters in a shared 3D space, while
UniMoCa~\cite{tan2026unimoca} represents human motion and
camera control through visual proxies.
PrevizWhiz~\cite{hu2026previzwhiz} connects rough 3D scene
authoring with generative models for video previsualization.
These methods enable explicit spatial and temporal control,
but rely on conditioning inputs that must be authored or
reconstructed separately.
We address this input preparation step by reconstructing
editable 3D blockouts from reference videos.
The recovered blockouts can be further edited and paired
with their source videos to provide aligned training data.
\section{Conclusion}
\label{sec:conclusion}

This work introduced \method for reconstructing editable 3D
blockouts from reference videos, together with \bench for
evaluation and \dataset for paired blockout--video data.
By grounding agentic modeling in metric observations and
explicit verification, \method improves reconstruction
quality and efficiency across different agents.
The resulting scenes support editing and provide structural
guidance for video generation.
Future work will improve reconstruction while reducing computational costs, and train video generators on paired data to better follow blockout-specified camera motion, spatial layout, and character blocking.
\section*{LLM Usage Statement}

Large language models (LLMs) were used in two ways in this work.
First, multiple LLM agents were employed as experimental systems for
agent-based 3D scene construction, including GPT-6.1 Sol, Claude Opus 5.5, DeepSeek-V4.1-Flash-1017, Claude Sonnet 5.5, and Kimi K3.1 Pro.
These models served as the agent backends evaluated under the Vanilla and proposed settings described in the paper; they were not used to generate scientific claims or final results without human verification. Second, ChatGPT Assist was used for editing and refining author-produced text, including grammar, clarity, and wording polish. All research ideas, experimental design, analyses, and final content were produced and verified by the authors, who take full responsibility for the paper.
{
    \small
    \bibliographystyle{ieeenat_fullname}
    \bibliography{main}
}


\end{document}